\documentclass[conf]{new-aiaa}
\usepackage[utf8]{inputenc}

\usepackage{graphicx}
\usepackage{amsmath}
\usepackage[version=4]{mhchem}
\usepackage{siunitx}
\usepackage{longtable,tabularx}

\usepackage[numbib]{tocbibind}
\usepackage{amsfonts}
\usepackage{amsmath}
\usepackage{xcolor}
\usepackage{soul}
\usepackage{placeins}
\usepackage{tablefootnote}
\usepackage{caption}
\usepackage{subcaption}
\usepackage[table]{xcolor}

\title{Online Action-Stacking Improves Reinforcement Learning Performance for Air Traffic Control}

\author{Ben Carvell\footnote{Co-Investigator, Project Bluebird, NATS} }
\affil{NATS, Whiteley, Hampshire, PO15 7FL, United Kingdom}
\author{George {De Ath}\footnote{Lecturer, Department of Computer Science, University of Exeter}}
\affil{University of Exeter, Exeter, England EX4 4QJ, United Kingdom}
\author{%
    Eseoghene Benjamin\footnote{Research Software Engineer, Project Bluebird, The Alan Turing Institute}
    and
    Richard Everson\footnote{Principal Investigator, Project Bluebird, The Alan Turing Institute and the University of Exeter}
}
\affil{The Alan Turing Institute, London, England NW1 2DB, United Kingdom}

\begin{document}

\maketitle

\begin{abstract}
We introduce \textit{online action-stacking}, an inference-time wrapper for reinforcement learning policies that produces realistic air traffic control commands while allowing training on a much smaller discrete action space. Policies are trained with simple incremental heading or level adjustments, together with an action-damping penalty that reduces instruction frequency and leads agents to issue commands in short bursts. At inference, online action-stacking compiles these bursts of primitive actions into domain-appropriate compound clearances. Using Proximal Policy Optimisation and the BluebirdDT digital twin platform, we train agents to navigate aircraft along lateral routes, manage climb and descent to target flight levels, and perform two-aircraft collision avoidance under a minimum separation constraint. In our lateral navigation experiments, action stacking greatly reduces the number of issued instructions relative to a damped baseline and achieves comparable performance to a policy trained with a 37-dimensional action space, despite operating with only five actions. These results indicate that online action-stacking helps bridge a key gap between standard reinforcement learning formulations and operational ATC requirements, and provides a simple mechanism for scaling to more complex control scenarios.
\end{abstract}

\section{Introduction}

\subsection{Background}

\lettrine{A}{utomation} of Air Traffic Control represents one of the most critical challenges in aviation. The International Air Transport Association (IATA) estimates that 5.2 billion passengers will fly in 2025~\citep{iata_stats}, with this figure projected to increase by up to 12.4 billion by 2050~\citep{icao_strategic}. The UK Government has matched the IATA ``Fly Net Zero'' resolution with their own ``Jet Zero'' commitment to net zero carbon emissions for aviation by 2050~\citep{jet_zero}. In this context, the need for advanced decision support systems that can revolutionise operational efficiency becomes increasingly urgent. Meeting these demands while maintaining stringent safety standards requires domain-relevant research that establishes robust foundations for Air Traffic Control Officer (ATCO) automation support.

ATCOs in the UK utilise a range of support systems to assist them in the controlling task, but the sophistication of these systems varies. The most advanced are predictive systems such as iFACTS~\citep{ifacts} and iTEC~\citep{itec}, which provide forecast predictions of aircraft behaviour driven by the Base of Aircraft DAta (BADA) model of aircraft performance developed by Eurocontrol~\citep{nuic_bada_2010}. These tools then compare these predictions to aid in identifying potential conflicts between aircraft, and to facilitate ``what-if'' probing of potential clearances to assess their ramifications. No suggestion of what clearance to issue is given, and responsibility for decisions and their safety rests entirely with the ATCO.

Research to move towards full decision automation in ATC has been ongoing since the work of Wesson et al. in the 1970s~\citep{wesson_planning_1977}, with a rich history of approaches studied and prototyped. Rules-based and heuristic approaches have been well documented, with the most mature example being ARGOS, a prototype system developed at Eurocontrol~\citep{argos}. Significant industry efforts have been made within the SESAR (Single European Sky ATM Research) programme, including AGENT~\citep{agent} on rules-based approaches to conflict resolution, and more recently HYPERSOLVER~\citep{hypersolver} using reinforcement learning to manage both flow management and tactical ATC simultaneously.

Reinforcement Learning (RL) has emerged as a promising approach to this challenge, with numerous studies exploring its application to ATC tasks~\citep{brittain_autonomous_2018, brittain_autonomous_2019, brittain_one_2021, wang_deep_2019, sui_conflict_2023, sui_tactical_2023, dalmau_air_2020, passerini_automating_2022}. Unlike traditional rules-based systems, RL offers the capacity to learn from experience and discover novel solutions to complex scenarios, as demonstrated in the groundbreaking work of DeepMind on AlphaStar and AlphaGo~\citep{alphastar, alphago}, OpenAI on Dota 2~\citep{berner2019dota}, and in other applications to domains such as robotics~\citep{smith2023demonstrating}. These characteristics make RL particularly suitable for the dynamic and nuanced decision-making required in air traffic control.

\subsection{Reinforcement Learning and PPO}

\begin{figure} [!h]
    \centering
    \captionsetup{width=1.0\linewidth}
    \includegraphics[width=0.5\textwidth]{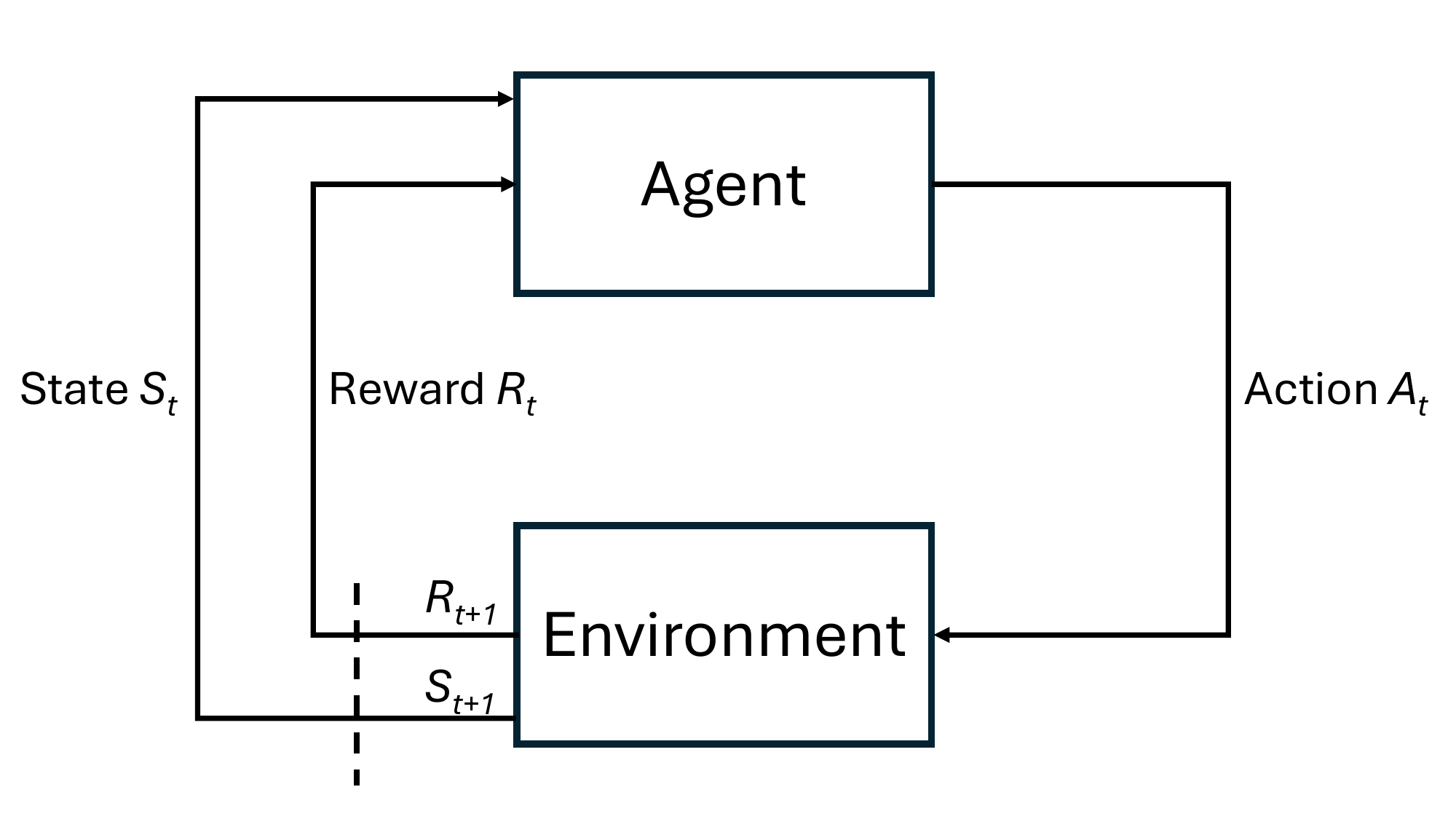}
    \caption{The interaction loop of agent with environment in a Markov Decision Process.}
    \label{fig:rl_diag} 
\end{figure}

In RL, an agent is trained to maximise the cumulative reward acquired by choosing actions to take within a given environment~\citep{sutton_reinforcement_2018}, with this interaction formulated as a Markov Decision Process~\citep{bellman_markovian_1957}. The agent interacts with the environment over a series of discrete time steps $\mathit{t} = 0, 1, 2, \dots$ and at each time step must select an action $\mathit{A}_t$ from a set of available actions $\mathcal{A}$ given the current observed state of the environment $\mathit{S}_t \in \mathcal{S}$. The representation of the state must be chosen such that Pr[$\mathit{S}_{t+1}\, | \, \mathit{S}_{t}, \mathit{A}_{t}$] = Pr[$\mathit{S}_{t+1}\, | \, \mathit{S}_1, \mathit{A}_1, \dots, \mathit{S}_{t}, \mathit{A}_{t}$] to satisfy the Markov property, i.e., the probability of transitioning to the successor state $\mathit{S}_{t+1}$ is independent of all previous states and actions except for the current state $\mathit{S}_t$ and action $\mathit{A}_t$. As shown in Figure~\ref{fig:rl_diag}, the environment then advances one time step, and the agent receives a reward $\mathit{R}_{t+1}$ and an updated observation of the state $\mathit{S}_{t+1}$, forming a tuple of state, action, the resulting reward, and the next state $\langle \mathcal{S}_t$, $\mathit{A}_t$, $\mathit{R}_{t+1}$, $\mathit{S}_{t+1} \rangle$.

By processing the trajectory of these tuples through time over many repeated training episodes, an agent can learn which actions to choose given a specific state to maximise the cumulative reward it receives. Agent behaviour is modulated through a discount factor $\gamma \in (0, 1]$, which controls how much an agent prefers to receive immediate reward rather than predicted future reward. A low value of $\gamma$ will result in a preference for immediate reward, whereas a high $\gamma$ will result in the potential sacrifice of immediate reward for the chance of higher total reward later. 
There are two primary methods for training agents, policy-based and value-based methods. Policy-based methods learn a policy $\pi$($\mathit{s}$) which maps any given state to a resultant action (or probability of choosing actions), while value-based methods learn a value function $\mathit{V}^\pi$($\mathit{s}$) which predicts the expected return (total reward from the current time step until the end of an episode) given a particular state when following a policy~\citep{sutton_reinforcement_2018}.

For this work, we use Proximal Policy Optimisation (PPO)~\citep{schulman_proximal_2017}, a state-of-the-art reinforcement learning method that combines elements of the two aforementioned families of methods, in a configuration known as actor-critic. In this configuration, both a policy $\pi$ and a value function $\mathit{V}^\pi$ are trained, with the value function then being used as the ``critic''. By recording the actual total reward accrued under a policy, and then comparing this with the expected return given by the value function, a measure of how much better or worse an agent is performing against prior experience is obtained, with this used to guide updates to the policy. Combined with an efficient method for limiting the magnitude of policy updates to prevent instability in training, PPO has become widely regarded as a best-in-class method, being used for DeepMind's Alphastar~\citep{alphastar} and for refining output in recent groundbreaking Large Language Models~\citep{ouyang_training_2022}.

\subsection{Prior Work: RL Formulations of Air Traffic Control}

The suitability of reinforcement learning for solving sequential decision-making problems under uncertainty has led to a range of applications to the controlling task. The work of \citet{brittain_autonomous_2019} stands out as one of the first of the new wave of publications that attempt to automate the ATC task via reinforcement learning. The authors formulated ATC as a multi-agent reinforcement learning problem, and use the Bluesky simulator~\citep{hoekstra_bluesky_2016} to create their environment. The work showcased the potency of RL as a viable approach for automating ATC. Despite its success, the authors employed a simplified scenario setup (relative to the full 3D ATC task) by restricting the available actions $\mathcal{A}$ of the agent to only increase, decrease, or hold the current speed of the aircraft. This action space is then used to solve a scenario where aircraft fly at the same level along a number of different routes, with the speed adjustments used to ensure they maintain a required separation of 3 nautical miles both in trail and against aircraft on different routes at crossing points.

The restriction of the environment design and action space is a common feature of other approaches in the literature which have sought to apply RL to elements of ATC. The work of Wang et al.~\citep{wang_deep_2019} also considers the lateral deconfliction element only, but introduces a continuous action space of heading adjustments rather than speed modifications. \citet{dalmau_air_2020} extend to a more fully featured lateral action space by including both speed and lateral navigation actions, but still omit the vertical element of control. The omission of vertical control is prevalent in the current literature. Vertical control is a critical element of the control task required to achieve cruising levels, descending for landing, to meet a myriad of coordination agreements, and to carry out deconfliction using the vertical dimension.

These formulations heavily simplify the reward design for training, as they do not consider the complex measures of success which apply when considering the full 3D control task. Judging separation only in the lateral plane allows for simplified forecasting of aircraft interactions without the concern of different levels and vertical profiles.

The objective of minimising the number of clearances per aircraft is also often omitted, which forms the more general challenge of action sparsity for RL as explored in a non-ATC context in the work of \citet{biedenkapp_temporl_2021}. Limiting the number of clearances is necessary for a range of operational reasons, including managing a pilot's workload, managing ATCO workload, and managing loading on the radio frequency used for issuing commands. Although some of these factors may be mitigated by future developments in technology, the naturally slow progress of technology in ATC due to safety challenges and difficulties in mandating aircraft equipage means that ignoring this key constraint severely impacts near-term relevance of research.

Some recent papers more closely represent the full control task. \citet{sui_conflict_2023} employ actions in both the lateral and vertical planes, but restrict the formulation to conflict detection and resolution between a pair of aircraft, disregarding the wider goals of the sector. The work of \citet{passerini_automating_2022} is well-formed from a domain perspective, with an action space covering vertical, lateral, and speed modifications, as well as utilising ATC scenario design that includes multiple aircraft and broader sector goals. The safety reward function in this work is simplified however, as only losses of separation are penalised, as opposed to incentivising a fail-safe method of operation as required in real-world operations. This significantly expands the viable solution space to decisions which would be judged extremely unsafe in the real world, again limiting the relevance of the results from the perspective of domain application. The Machine Basic Training framework \cite{bluebird_validation} proposes a clear set of performance objectives for early maturity ATC agents, adapted from a real ATCO training curriculum. This framework makes clear that even for early maturity formulations of ATC automation a broad action space and stringent safety definitions are essential.

In summary, the majority of the current literature on RL applications for ATC presents a restricted problem statement, obscuring the challenges that exist when attempting to utilise RL for decision-making automation in this domain. Issues of variable state space are addressed, but variable and large action spaces, action sparsity, domain-accurate measures of safety, and the discipline of vertical control are under-represented or omitted entirely. This presents an opportunity for work that addresses these challenges in combination and develops techniques for training agents under these multiple challenging conditions.

\subsection{Motivation and Contribution}

A key objective of air traffic control is to issue the fewest clearances necessary to effectively control aircraft. This minimises the workload for pilots and reduces congestion on sector radio frequencies, ensuring that all required clearances can be issued in a timely manner. This operational requirement conflicts with common RL formulations for ATC, where headings and levels are modified through small incremental adjustments, leading to numerous commands to achieve what operationally would be a single instruction. In practice, an Air Traffic Control Officer (ATCO) would issue one directive ("\textit{turn left 30 degrees}") rather than three consecutive commands ("\textit{turn left 10 degrees}"), and would only provide instructions when necessary rather than constantly fine-tuning aircraft paths. Although state information is updated regularly (one radar refresh every 6 seconds in the UK) in the same manner as turn-taking games, these operational constraints distinguish ATC from other high-profile RL applications such as Go, Chess, or StarCraft. Success in ATC depends not only on the final outcome but on the efficiency and manner of execution.

In this work, we introduce \textit{online action-stacking}, an online policy-driven approach that enables agents trained with limited discrete actions to produce comprehensive ATC commands while minimising the total number of instructions issued. This technique conceptually relates to the \textit{options} framework of \citet{sutton:macroactions:1999}, but is specifically tailored for controller-style directives. However, in contrast to options, FiGAR~\citep{figar}, and TempoRL~\citep{biedenkapp_temporl_2021}, we do not change the training objective or introduce semi-MDPs. Instead, we wrap a standard PPO policy with an inference-time procedure that compiles primitive commands into macro-instructions while preserving the underlying MDP structure. By addressing the action frequency challenge, action stacking significantly simplifies the training of agents to perform fundamental ATC behaviours and provides a critical foundation for more domain-accurate implementations.

\section{Methodology}
\subsection{Scenario Specification}

Operational ATC presents a complex and multi-faceted task, requiring ATCOs to maintain safety between aircraft while providing efficient lateral and vertical profiles, and meeting transfer constraints (known as coordination) between sectors. Here, we describe how we composed scenarios to train and test aspects of this challenge.

\subsubsection{ATC Problem Statement}
We designed scenarios to train and evaluate agent competence in four fundamental ATC behaviours:
\begin{itemize}
    \item Navigate aircraft along a route through a sector using heading instructions, ensuring that it exits at the designated route point
    \item Navigate aircraft to their coordinated exit level with the next sector by issuing vertical instructions
    \item Deconflict aircraft in the lateral plane, ensuring that aircraft trajectories always maintain the $5$ nautical mile separation standard even if no further instructions are issued
    \item Minimise the number of instructions issued while effectively maintaining control
\end{itemize}
These behaviours align directly with the requirements of CAP493 for tactical controllers to provide safe, orderly, and expeditious control~\citep{CAP493}. The scenarios were run for 300 steps (each step representing 6 seconds of simulation time), which provided sufficient time for aircraft to transit the airspace. Scenarios were generated using either one or two aircraft, depending on the use case. We trained lateral navigation and avoidance with two aircraft, and vertical navigation with one aircraft. Training and test scenarios were generated stochastically with the same process, with random variations in start position and route as described in Section \ref{sec:Aircraft}.

\subsubsection{Airspace}\label{sec:Airspace}
\begin{figure}[!htb]
    \centering
    \includegraphics[width=0.63\textwidth]{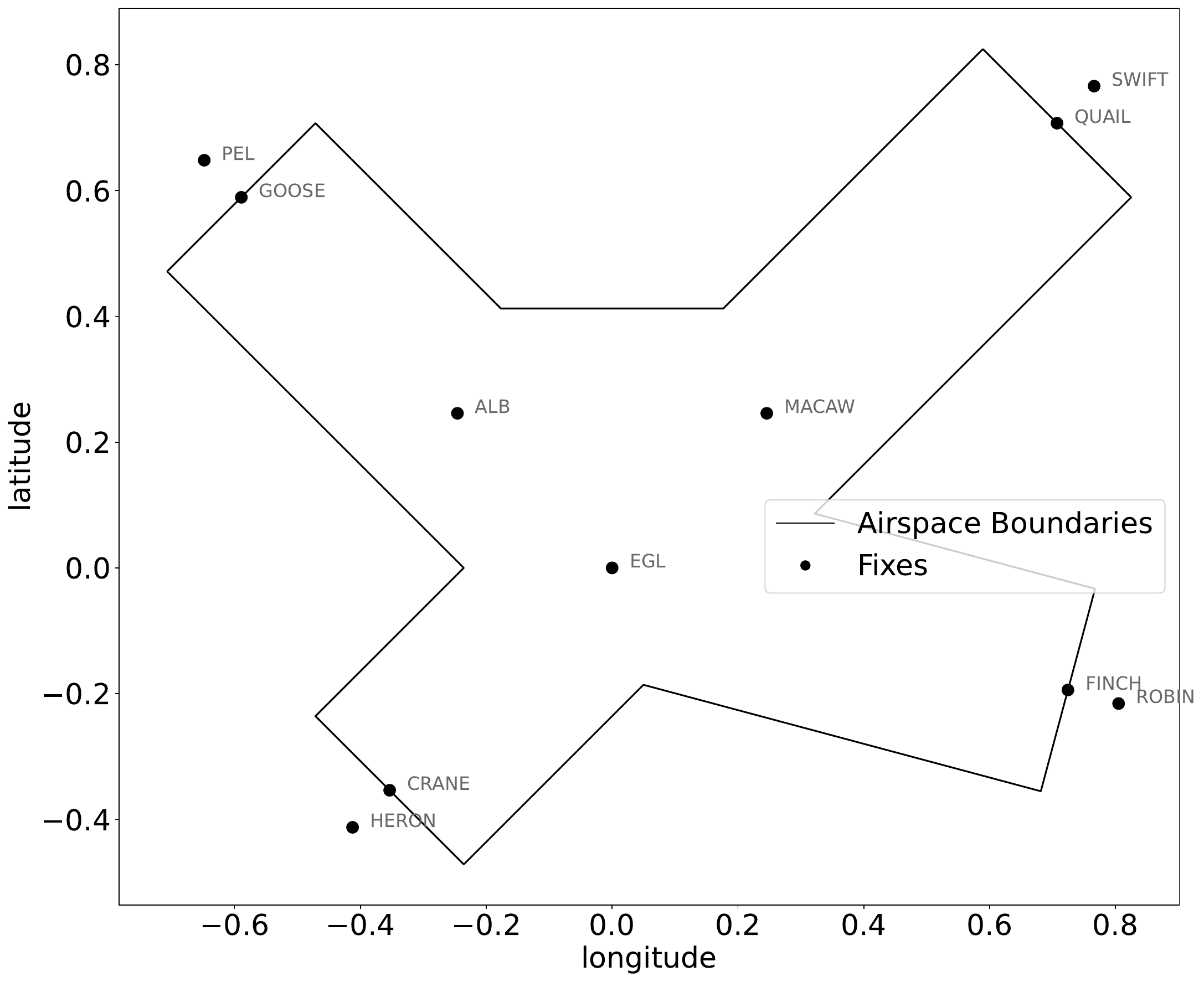}
    \caption{The ``X-Plus Sector'', an artificial representation of en-route airspace with 20 nm airways and named navigation fixes.}
    \label{fig:airspace} 
\end{figure}

\begin{figure}[!htb]
    \centering
    \includegraphics[width=0.85\textwidth]{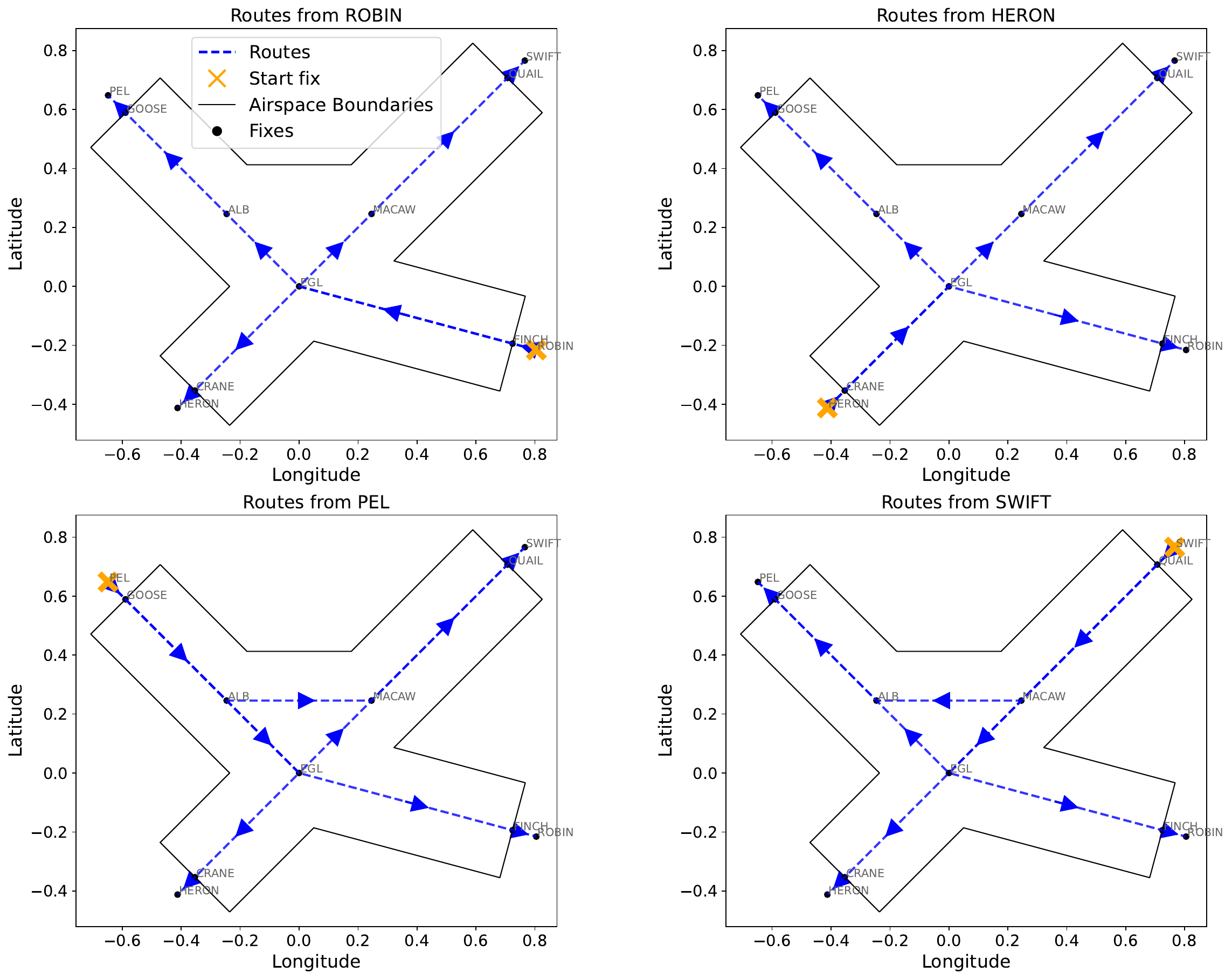}
    \caption{Viable routes through the X-Plus sector from each starting point.}
    \label{fig:airspace_with_routes} 
\end{figure}

All scenarios within this work use a test airspace formulated as part of \textbf{\textit{Project Bluebird}}\footnote{Project Bluebird is an EPSRC Prosperity Partnership between NATS, The Alan Turing Institute and The University of Exeter.}~\citep{bluebird}, the ``X-Plus Sector'', as shown in Figure ~\ref{fig:airspace}. This airspace features airways that are 20 nautical miles wide and is constrained to fit within a 120 nm bounding box, being representative of a standard en-route sector configuration. The airspace contains navigational fixes that delineate routes that aircraft can take, with the outermost fixes acting as ``seed points'' from which to create aircraft. It is constructed to provide a challenge to navigate aircraft and to present potential aircraft interactions at route intersections. The viable routes through the airspace are shown in Figure \ref{fig:airspace_with_routes}.

\subsubsection{Aircraft}\label{sec:Aircraft}

For lateral control scenarios, two aircraft were created with the parameters shown in Table \ref{tab:lat_ac_params}. For vertical control scenarios, a single aircraft was created with the parameters shown in Table \ref{tab:vert_ac_params}.

\begin{table}
\centering
\caption{Aircraft parameters for lateral control scenarios}
\begin{tabular}{l | l} 

 Parameter & Value\\ [0.5ex] 
 \hline
 Aircraft Type & Boeing 757-300\\ 

 Route & Random choice from viable sector routes${^*}$\\

 Level & Flight Level 300 (30,000 feet)\\

 Start Position & Randomly positioned within a 4nm radius circle centred on the first route fix\\

 Heading & Parallel to initial route leg\\ [1ex]

\end{tabular}
\begin{center}
    *Route start points were deconflicted to prevent conflicts already being present at scenario initialisation.
\end{center}
\label{tab:lat_ac_params}
\end{table}

\begin{table}
\centering
\caption{Aircraft parameters for vertical control scenarios}
\begin{tabular}{l | l} 

 Parameter & Value\\ [0.5ex] 
 \hline
 Aircraft Type & Boeing 757-300\\ 

 Route & Random choice from viable sector routes${^*}$\\

 Initial Level & Random choice between Flight Level 100 and Flight Level 300\\

 Target Level & Random choice between Flight Level 100 and Flight Level 300\\

 Start Position & At first route fix\\

 Heading & Following own navigation along route\\ [1ex]

\end{tabular}
\label{tab:vert_ac_params}
\end{table}

\subsection{Problem Formulation}
We utilise the BluebirdDT~\citep{bluebird_dt} digital twin platform\footnote{This platform is being developed as part of Project Bluebird~\citep{bluebird}, and will form part of an upcoming open source release.} for our simulations. The performance of the aircraft is modelled using a simplified implementation of the BADA physics-based model~\citep{nuic_bada_2010}. The simulator updates aircraft states at six-second intervals, matching operational radar update rates. 

This problem setup naturally aligns with the structure of a Markov Decision Process. At any time $\mathit{t}$ the current state $\mathit{S}_t$ of the simulator together with the current action $\mathit{A}_t$ provides the necessary information to predict, with uncertainty and without relying on history, the next state of the simulator, embodying the Markov property. At each time step, the agent selects an action $\mathit{A}_t \in \mathcal{A}$ to modulate the behaviour of an aircraft within the environment, determined by a policy $\pi$ (\textit{no action} is part of this action set). Finally, after the execution of an action $\mathit{A}_t$, the updated environment state $\mathcal{S}_{t+1}$ may be analysed to calculate a reward $\mathit{R}_{t+1}$, judging the evolution of the environment arising from action $\mathit{A}_t$ against the objectives of good tactical control. We utilise a centralised formulation of RL (as opposed to multi-agent), and as such our state vectors for multiple aircraft are concatenated into a single vector, with a similar approach applying to the action space. The details of this formulation are given below.

\subsubsection{State}
As described by \citet{asadi_state_2004}, reinforcement learning configurations train faster when redundant variables are removed from the state space while retaining all information necessary for predicting the next state. To maximise future scalability, we aim for these simple initial examples to use domain knowledge to select only essential variables for the state space, improving training speed while retaining the ability to train effectively.

The state vector $\mathcal{S}$ for a single aircraft in lateral control scenarios is formulated as shown in Table \ref{tab:lat_state}.

\begin{table}
\centering
\caption{State variables for lateral control scenarios}
\begin{tabular}{p{1.2cm} | p{10cm} | p{1.6cm} | p{2cm}} 
 Feature & Definition & Range & Units\\ [0.5ex] 
 \hline
 $\theta_{\it{f}}$ & Relative turn to next fix from cleared heading & [-180, 180] & degrees\\ 

 $\theta_{\it{sf}}$ & Relative turn to subsequent fix (next fix plus one) from cleared heading & [-180, 180] & degrees\\ 

 $\mathit{d_{c}}$ & Clipped distance from route centreline & [-100, 100] & nautical miles\\

 $\mathit{\Delta_{t}}$ & Clipped time since last action & [0, 60] & seconds\\ [1ex]

\end{tabular}
\label{tab:lat_state}
\end{table}
The turn to the subsequent fix is included to allow for effective manoeuvring when avoiding other aircraft by giving more awareness of the exit target.
The term ``cleared heading'' refers to the heading that the aircraft has been instructed to follow. The distance from the route centreline is clipped to a range of 100nm because all values in the state vector require defined bounds, and 100nm adequately bounds the ``area of interest'' for the scenario. Behaviours outside these bounds are not relevant to the study as they represent major excursions away from the airspace and the route. The time since the last action is clipped to a maximum of 60 seconds, as this matches the range over which we penalise taking another action as shown later in Equation~\ref{sparse_action}.

When running 2-aircraft scenarios, this observation space is doubled, with one set of values per aircraft. When running 2-aircraft scenarios with avoidance, two additional variables are included as shown in Table \ref{tab:avoid_state}.
\begin{table}
\caption{Extra state variables for lateral avoidance scenarios}
\centering
\begin{tabular}{p{1.2cm} | p{10cm} | p{1.6cm} | p{2cm}} 
 Feature & Definition & Range & Units\\ [0.5ex] 
 \hline

 $\theta_{\it{1,2}}$ & Relative turn from aircraft 1 to aircraft 2 & [-180, 180] & degrees\\ 

 $\mathit{d_{1,2}}$ & Distance between aircraft 1 and aircraft 2 & [0, 150] & nautical miles\\ [1ex]

\end{tabular}
\label{tab:avoid_state}
\end{table}
A full state vector for two aircraft lateral avoidance would then be: 
\begin{equation}
    \mathcal{S}_t = (\theta_{\it{f1}}, \theta_{\it{fs1}}, \mathit{d_{c1}}, \mathit{\Delta_{t1}}, \theta_{\it{f2}}, \theta_{\it{fs2}}, \mathit{d_{c2}}, \mathit{\Delta_{t2}}, \theta_{\it{1,2}}, \mathit{d_{1,2}}) 
\end{equation}

Finally, when training simple single aircraft vertical control scenarios, we provide the state vector with the elements shown in Table \ref{tab:vert_state}

\begin{table}
\centering
\caption{State variables for vertical control scenarios}
\begin{tabular}{p{1.2cm} | p{10cm} | p{1.6cm} | p{2cm}} 
 Feature & Definition & Range & Units\\ [0.5ex] 
 \hline
 $\mathit{d_{xfl}}$ & Clipped difference between distance to exit fix and distance required to climb/descend from current level to exit level at a set rate of 2000 feet per minute & [-100, 100] & nautical miles\\ 

 $\mathit{FL_{sx}}$ & Clipped difference between \textbf{s}elected flight level and e\textbf{x}it flight level & [-200, 200] & flight levels\\ 

 $\mathit{FL_{nx}}$ & Clipped difference between e\textbf{n}try (initial) flight level and e\textbf{x}it flight level & [-200, 200] & flight levels\\ 

 $\mathit{\Delta_{t}}$ & Clipped time since last action & [0, 10] & seconds\\ [1ex]

\end{tabular}
\label{tab:vert_state}
\end{table}

The state values are normalised to lie in the range [0, 1] using the ranges defined in the above tables.

\subsubsection{Action}\label{sec:Action}
Reduced action spaces are also known to result in faster and more reliable policy training~\citep{kanervisto_action_2020}.
The real action space for operational ATC is extremely large. Just for lateral vectoring commands, a full $360^\circ$ in $5^\circ$ increments can be issued (72 commands), as well as a range of relative turns, left and right at least $30^\circ$ in $5^\circ$ increments (12 commands). Add to this route navigation instructions, speed, and vertical instructions (between approximately flight level 70 and 450 in ten level increments for en-route), and the per-aircraft action space easily exceeds 100 actions, and likely approaches 200. For our scenarios, we train with incremental instructions in the lateral and vertical planes and then use our novel technique of online action-stacking to compile them into operationally realistic commands.

Our action space for a single aircraft is:
\begin{equation}
        \mathcal{A} = \{\phi, \mathit{h_{-}}, \mathit{h_{+}}\} 
\end{equation}
where $\phi$ is no action, $\mathit{h_{-}}$ is \textit{turn left 10 degrees}, and $\mathit{h_{+}}$ is \textit{turn right 10 degrees}. A turn size of 10 degrees was chosen to give a reasonable fidelity of control while ensuring a turn takes longer than one step of the simulator to execute.

For two aircraft scenarios, the action space is duplicated for each aircraft: 
\begin{equation}
    \mathcal{A} = \{\phi, \mathit{h^1_{-}}, \mathit{h^1_{+}}, \mathit{h^2_{-}}, \mathit{h^2_{+}}\}
    \label{eq:two_ac_act_space}
\end{equation}
\newpage
When we train with a large lateral action space as a reference implementation, the action space for a single aircraft is:
\begin{equation}
        \mathcal{A} = \{\phi, \mathit{h_{-10}}, \mathit{h_{-20}}, \mathit{h_{-30}}, \mathit{h_{-40}}, \mathit{h_{-50}}, \mathit{h_{-60}}, \mathit{h_{-70}}, \mathit{h_{-80}}, \mathit{h_{-90}}, \mathit{h_{+10}}, \mathit{h_{+20}}, \mathit{h_{+30}}, \mathit{h_{+40}}, \mathit{h_{+50}}, \mathit{h_{+60}}, \mathit{h_{+70}}, \mathit{h_{+80}}, \mathit{h_{+90}}\} 
        \label{2_ac_big_act_space}
\end{equation}
where $\phi$ is no action, $\mathit{h_{-10}}$ is \textit{turn left 10 degrees}, $\mathit{h_{-20}}$ is \textit{turn left 20 degrees}... in total giving a range of up to $90^\circ$ left or right in increments of $10^\circ$. In a two aircraft scenario, the heading actions are duplicated per aircraft, giving a total of 37 actions.

For vertical control scenarios, the action space for a single aircraft is:
\begin{equation}
        \mathcal{A} = \{\phi, \mathit{FL_{-}}, \mathit{FL_{+}}\} 
\end{equation}
where $\phi$ is no action, $\mathit{FL_{-}}$ is \textit{descend 10 flight levels}, and $\mathit{FL_{+}}$ is \textit{climb 10 flight levels}.

\subsubsection{Reward}\label{sec:Reward}

On each step of a scenario, we use multiple reward functions, formulated to provide negative reward for undesirable behaviours, which, when taken together as a weighted sum, form the total reward to be maximised. Deviations from the specified aircraft route are discouraged through the \textit{centreline distance reward}: 
\begin{equation}
    \mathit{r_{c}} = \exp \left( - (d_c/\lambda_c )^2 \right) - 1
    \label{centreline}
\end{equation}
where $\mathit{d_{c}}$ is the distance of the aircraft from the centreline of the route and $\lambda_c$ is a scaling factor. Using $\lambda_c = 6$ creates a gradual penalty near the centreline that increases sharply as aircraft approach the sector boundary ($10$nm from the centreline). This design reflects operational priorities: minor route deviations are acceptable, while potential airspace excursions that present safety risks are strongly discouraged. 

The \textit{action damping reward} is designed to limit the number of clearances issued:
\begin{equation}
  \mathit{r_{a}} =
    \begin{cases}
      \frac{n_s}{n_{max}} - 1 & \text{if } n_s < n_{max}\\
      0 & \text{otherwise},
    \end{cases}       
    \label{sparse_action}
\end{equation}
where $\mathit{n_s}$ is the number of time steps since the last action, and $\mathit{n_{max}}$ defines the threshold after which the negative reward ceases. For this experiment, $\mathit{n_{max}} = 10$, which is equivalent to one minute between clearances. This reward strongly discourages issuing commands too frequently, matching operational practice. In combination with the navigation and safety rewards, it leads policies to avoid small corrective actions and instead issue clustered sequences of instructions when larger manoeuvres are required. The slow decay of the negative reward following an action directly incentivises bursts of actions, as issuing actions consecutively with no gaps minimises the accrued negative reward. These emergent bursts of primitive actions are naturally amenable to online action-stacking.

The safe separation of aircraft is accomplished via the \textit{safety reward}:
\begin{equation}
  \mathit{r_{s}} =
    \begin{cases}
      \mathit{\left(\frac{d_{1,2}}{d_{max}} - 1\right)e^{-\left(\frac{d_{s}}{\lambda_s}\right)^2}} & \text{if } d_{1,2} < d_{max}\\
      0 & \text{otherwise}
    \end{cases}
    \label{eq:safety_reward}
\end{equation}
where $\mathit{d_{1,2}}$ is the current distance between the two aircraft, $\mathit{d_s}$ is the minimum distance predicted between the two aircraft when predicted forward in time for a path length of $\mathit{d_{max}}$, and $\lambda_s$ is a scaling factor. This formulation penalises states in which the projected minimum separation between the two aircraft becomes small or violates the 5 nm standard. The exponential term provides a small penalty when the predicted separation is comfortably above 5 nm, with a strong gradient towards the largest penalty as the projected separation approaches or falls below the required minimum. This penalty is linearly scaled by the current distance between the aircraft up to a maximum distance $\mathit{d_{max}}$, reflecting the increasing urgency of safety issues as aircraft get closer together. We set $\mathit{d_{max}} = 150$ to approximately match the size of the X-Plus sector, and $\lambda_s = 5$ to give the desired profile around the minimum separation standard of 5 nm, strongly discouraging safety violations.

For vertical scenarios, we implement a \textit{simple vertical reward}:
\begin{equation}
    \mathit{r_{v}} = \exp \left(\frac{- |\Delta_{FL}|}{\lambda_v} \right) - 1,
    \label{eq:vertical_reward}
\end{equation}
where $\Delta_{FL}$ is the difference between an aircraft's selected flight level and its exit flight level, and $\lambda_v$ is a scaling factor set to $40$ flight levels to give an effective gradient around the expected range of $\pm$200 flight levels.
\newpage

At the end of a scenario, a further set of terminal rewards is evaluated and added to the cumulative reward to give the scenario's final reward. These terminal rewards apply to both lateral and vertical scenarios. The \textit{terminal reward set} consists of a flat bonus reward dependent on agent success against 3 criteria:
\begin{enumerate}
    \item Navigation to exit at the correct level
    \item Remaining within the lateral and vertical sector bounds
    \item Issuing fewer than 30 actions per aircraft 
\end{enumerate}
A bonus reward of 5 per satisfied criterion is provided, and a further bonus of 20 when all criteria are satisfied. These rewards help to further aid convergence towards desired behaviours through providing an additional coarse success signal alongside the shaped reward functions.

These rewards are combined as weighted sum, dependent on scenario configuration. Table \ref{tab:reward_weights} shows the configurations used. These reward weights were obtained through iterative testing of a range of configurations whilst monitoring for success through the criteria described in the terminal reward set.

\begin{table}
\centering
\caption{Reward configuration weightings}
\begin{tabular}{l | c c c c c}
Configuration & Centreline Distance & Action Damping & Safety & Vertical & Terminal Set \\[0.5ex]
\hline
Lateral Navigation without damping  & 1.0 & \cellcolor{lightgray}0.0  & \cellcolor{lightgray}0.0 & \cellcolor{lightgray}0.0 & 1.0\rlap{\textsuperscript{*}} \\
Lateral Navigation                  & 1.0 & 0.25 & \cellcolor{lightgray}0.0 & \cellcolor{lightgray}0.0 & 1.0 \\
Vertical Navigation                 & \cellcolor{lightgray}0.0 & 0.5  & \cellcolor{lightgray}0.0 & 1.0 & 1.0 \\
Lateral Navigation and Avoidance    & 1.0 & 0.3  & 2.0 & \cellcolor{lightgray}0.0 & 1.0 \\[1ex]
\end{tabular}
\begin{center}
    *The action incentive was omitted from the terminal set in this configuration
\end{center}
\label{tab:reward_weights}
\end{table}

In summary, for lateral navigation scenarios, we train with the \textit{centreline distance reward}, the \textit{action damping reward}, and the \textit{terminal reward set}. For lateral navigation and avoidance scenarios, we train with the \textit{centreline distance reward}, the \textit{action damping reward}, the \textit{safety reward}, and the \textit{terminal reward set}. For vertical navigation scenarios, we train with the \textit{simple vertical reward}, the \textit{action damping reward}, and the \textit{terminal reward set}.

\subsection{Agent Training Configuration}

Our policy is implemented as a fully-connected neural network with $2$ hidden layers of $64$ neurons each, using ReLU activation functions. We configure our PPO implementation based in part on parameters from \citet{brittain_autonomous_2019}, with learning rate $\mathit{l_r} = 0.0001$, discount factor $\mathit{\gamma} = 0.99$, and entropy coefficient $\mathit{\beta} = 0.01$. The scenario episodes were terminated after $300$ time steps ($30$ minutes of simulated time). Scenarios were trained for a total of 2 million time steps. When training lateral avoidance policies, we adopt a simple curriculum-based approach: the policy is first pre-trained for 2 million steps while omitting the \textit{safety reward} in order to learn simple navigation to exit. The \textit{safety reward} is then included, and the policy is trained for a further 30 million time steps to account for the increase in task difficulty.

\subsection{Online Action-Stacking}
Online Action-Stacking takes advantage of the action-damping penalty described in Equation~\ref{sparse_action}. The penalty's slow decay encourages \textit{bursts} of actions, i.e., sequences of consecutive commands that achieve a desired heading change in a single decision point, minimising the cumulative penalty.

To implement action stacking, we repeatedly query the policy while it continues to issue commands for the same aircraft; we accumulate identical $10$ degree increments (left or right) into a single macro-command, and stop stacking when the policy outputs ‘no action’ or an action on the other aircraft. We keep the underlying training MDP unchanged. At inference, we perform several policy queries at a single physical time step, updating the commanded heading while the environment state (position, velocity) is held fixed. Each intermediate query is conditioned on a fully observed state that includes the updated commanded heading, so the Markov structure of the underlying process is preserved.

Unlike hierarchical or temporally abstract RL approaches that rely on predefined macro-actions or structured decompositions~\citep{sutton:macroactions:1999, asadi_state_2004}, online action-stacking dynamically compiles primitive actions based on the policy's output at runtime. This offers significant computational advantages by maintaining a minimal action space of just three discrete actions per aircraft during training, rather than the tens of possible heading commands that would otherwise be needed~\citep{kanervisto_action_2020}. The approach is particularly beneficial for ATC applications where specific heading values (rather than just relative changes) are operationally required~\citep{dalmau_air_2020}, and where coordination between multiple aircraft requires precise timing of commands~\citep{brittain_autonomous_2019}.

Additionally, by compounding actions without advancing the simulation time, we avoid the additional partial observability introduced by macro-actions that execute over multiple unobserved time steps. In traditional temporal abstraction approaches, when a macro-action executes over multiple time steps, the environment continues to evolve, but the agent cannot observe intermediate states or adjust decisions until the macro-action completes. This creates significant partial observability, as the environment state may change unpredictably during execution. Our approach avoids this problem by compiling multiple primitive actions at a single decision point before the environment advances.

\section{Results and Discussion}

To effectively test and demonstrate our contribution, we train a broad set of experimental policies. These are presented and examined in the following subsections.

\subsection{Navigation without Action Damping}

\begin{figure}
\centering
\begin{minipage}[t]{.6\textwidth}
  \centering
  \captionsetup{width=.8\linewidth}
  \includegraphics[width=.9\linewidth]{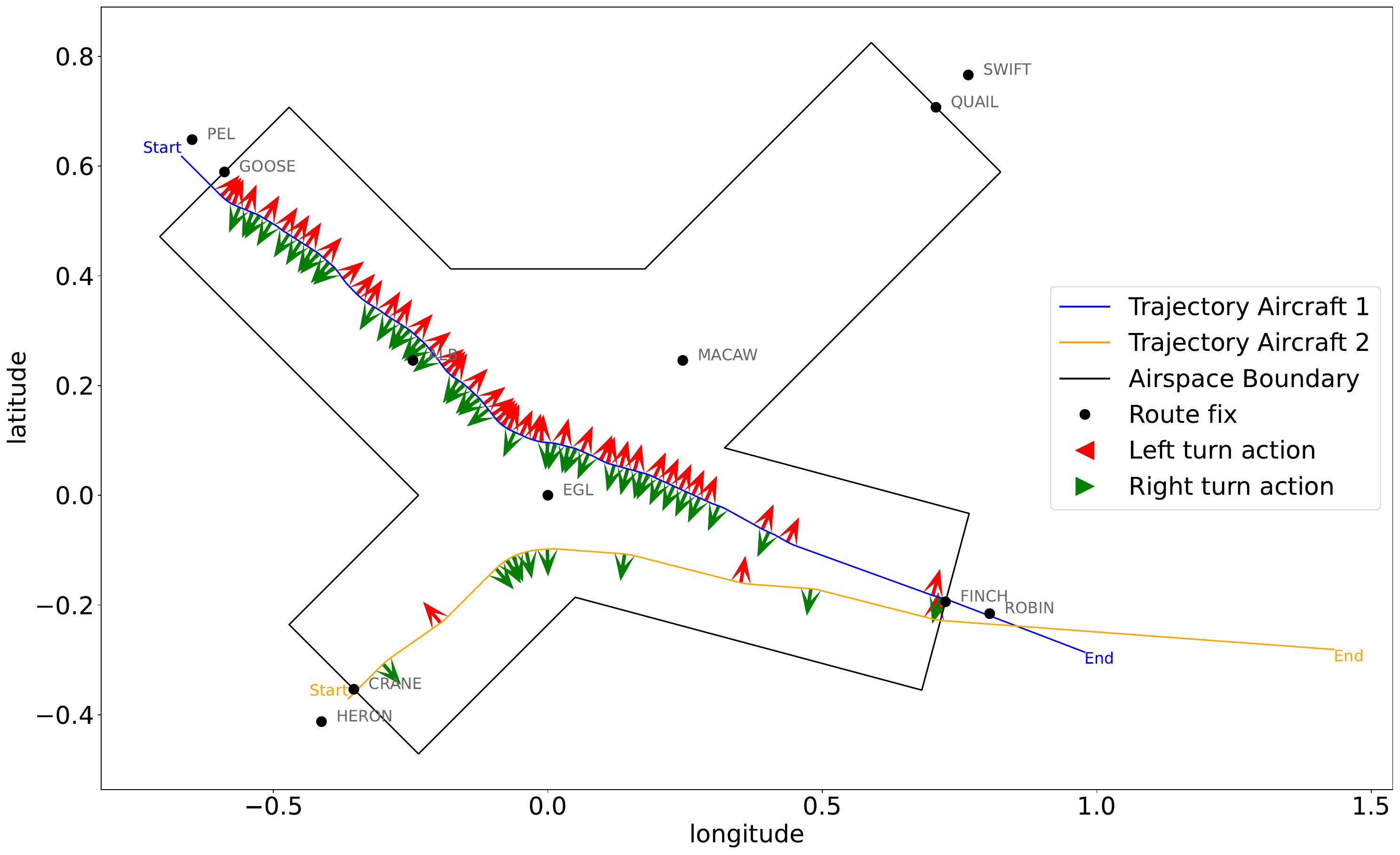}
  \vspace*{\fill}
  \captionof{figure}{Policy controlling 2 aircraft through the X-Plus sector with no action damping penalty and no action stacking. A total of 92 separate actions were issued.}
  \label{fig:undamped_example}
\end{minipage}%
\begin{minipage}[t]{.4\textwidth}
  \centering
  \captionsetup{width=.8\linewidth}
  \includegraphics[width=.9\linewidth]{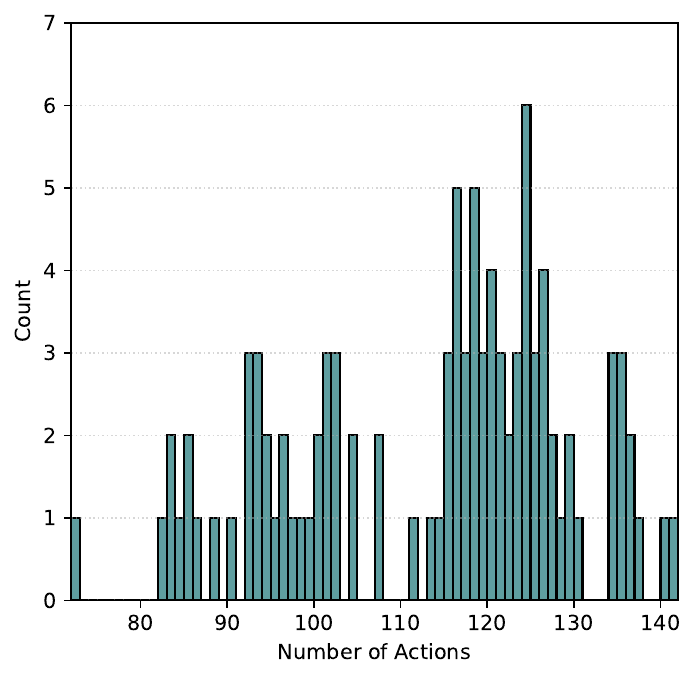} 
  \vspace*{\fill}
  \captionof{figure}{Number of actions taken per episode for a sample of $100$ episodes using an undamped lateral navigation policy for 2 aircraft navigation. A mean of 113.0 actions with standard deviation of 15.9 is observed.}
  \label{fig:undamped_act_counts}
\end{minipage}
\end{figure}

Figure~\ref{fig:undamped_example} shows the behaviour of a policy trained for 2 million time steps using only the lateral centreline distance reward described in Equation~\ref{centreline}, and the terminal reward set for successful navigation to the route exit and lack of excursions. Without a penalty for taking multiple actions, the policy issues an action almost every time step, resulting in a very high total number of actions per episode. Figure~\ref{fig:undamped_act_counts} shows the number of actions taken per episode for a sample of 100 episodes, with the policy issuing a mean of 113 actions. In addition to a high frequency of actions, this policy also demonstrates frequent oscillations between left and right control actions as it attempts to maintain the optimal distance from the centreline of the route, as can be seen in Figure \ref{fig:undamped_example}. This behaviour is a blocker to applying \textit{online action-stacking}, as we require repeated bursts of the same action.

\subsection{Damped Navigation and Action Stacking}

\begin{figure}
\centering
\begin{minipage}{.5\textwidth}
  \centering
  \captionsetup{width=.8\linewidth}
  \includegraphics[width=.9\linewidth]{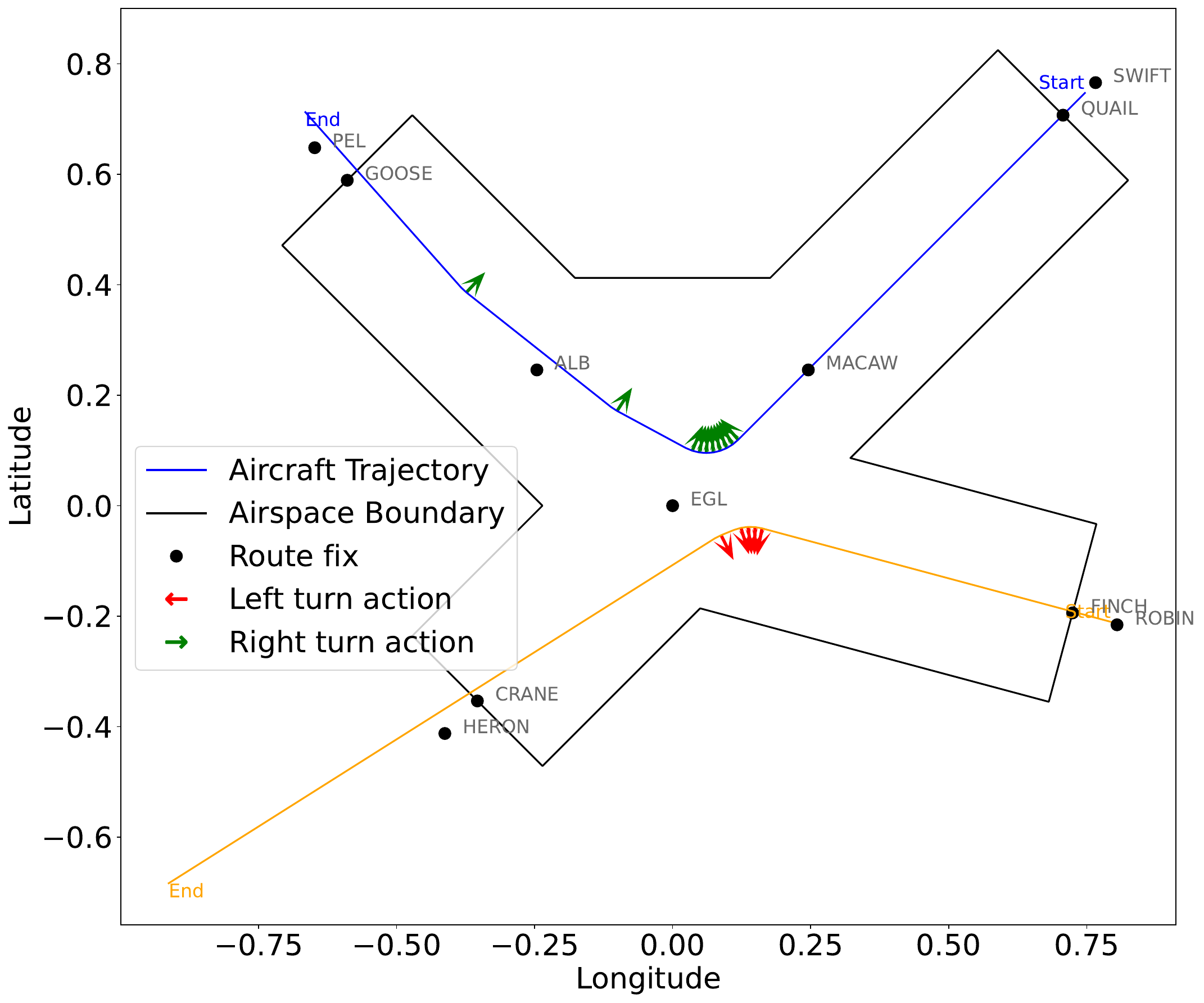}
  \captionof{figure}{Policy controlling 2 aircraft through the X-Plus sector with action damping penalty and no action stacking. The policy issues a total of 15 separate actions.}
  \label{fig:2_ac_damped_lat_example}
\end{minipage}%
\begin{minipage}{.5\textwidth}
  \centering
  \captionsetup{width=.8\linewidth}
  \includegraphics[width=.9\linewidth]{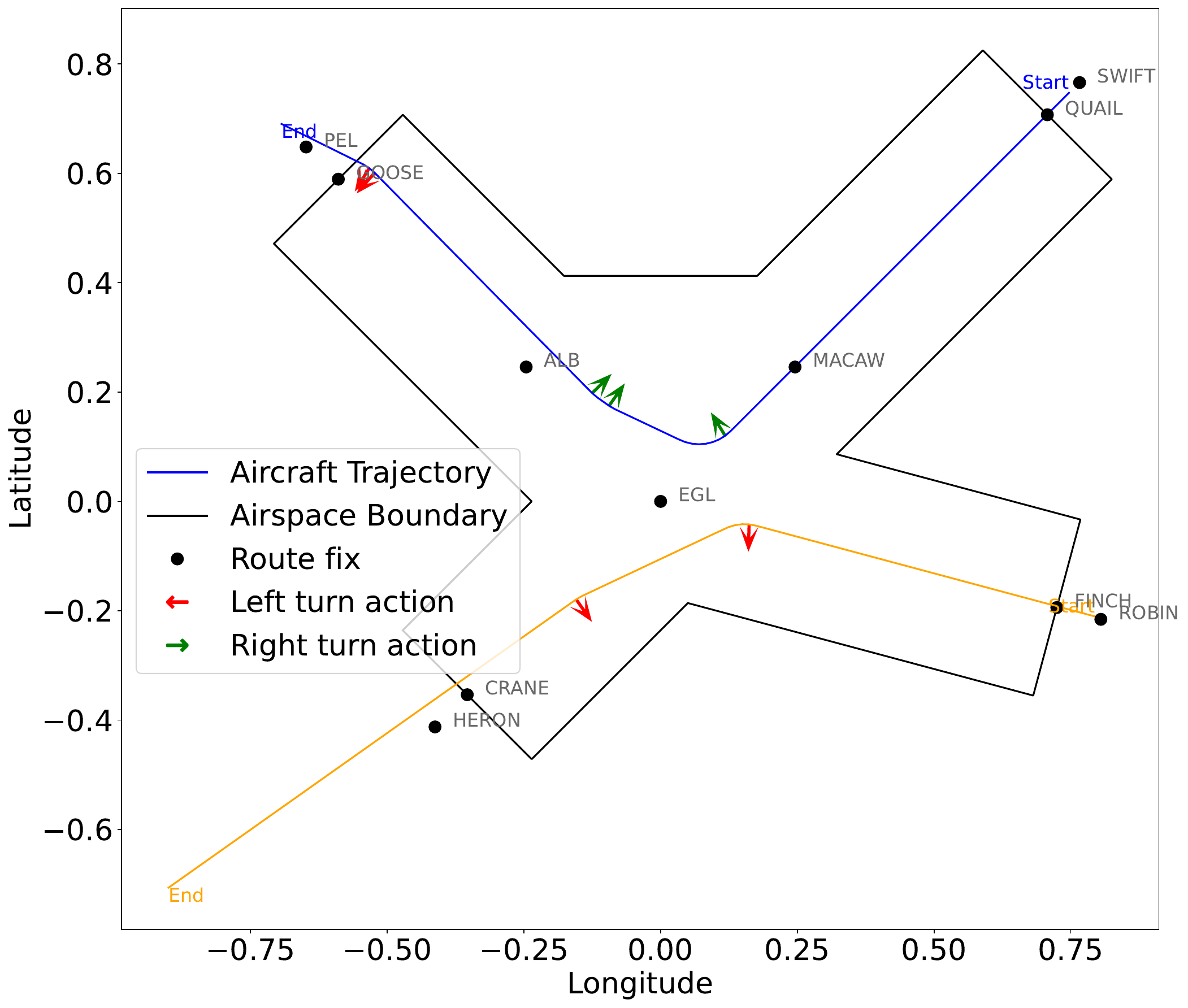}
  \captionof{figure}{Policy controlling 2 aircraft through the X-Plus sector with action damping penalty and action stacking. The policy issues a total of 7 separate actions.}
  \label{fig:2_ac_stacked_lat_example}
\end{minipage}
\end{figure}

\begin{figure}
\centering
\begin{minipage}[t]{.5\textwidth}
  \centering
  \captionsetup{width=.8\linewidth}
  \includegraphics[width=.9\linewidth]{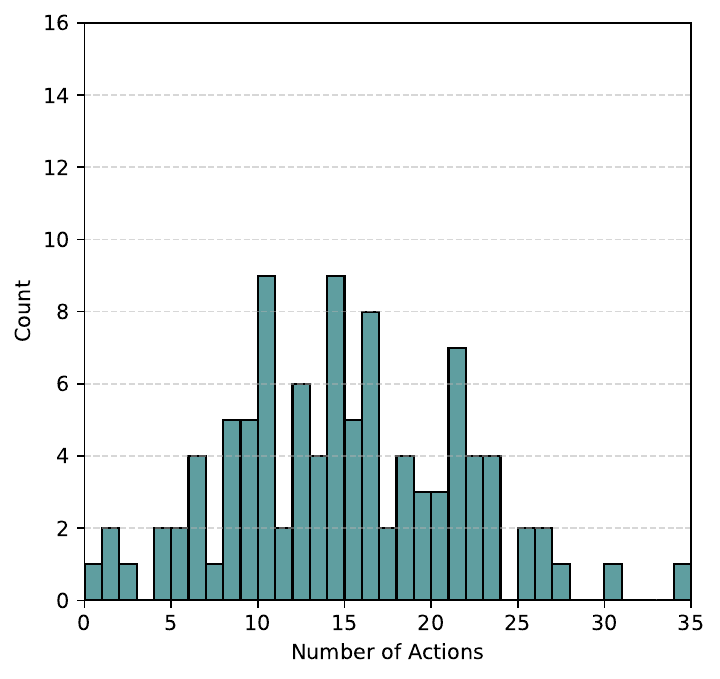}
  \captionof{figure}{Number of actions taken per episode for a sample of 100 episodes using a damped lateral navigation policy for 2 aircraft. A mean of 14.5 actions with standard deviation 6.6 is observed.}
  \label{fig:2_ac_lat_damp_multi_run}
\end{minipage}%
\begin{minipage}[t]{.5\textwidth}
  \centering
  \captionsetup{width=.8\linewidth}
  \includegraphics[width=.9\linewidth]{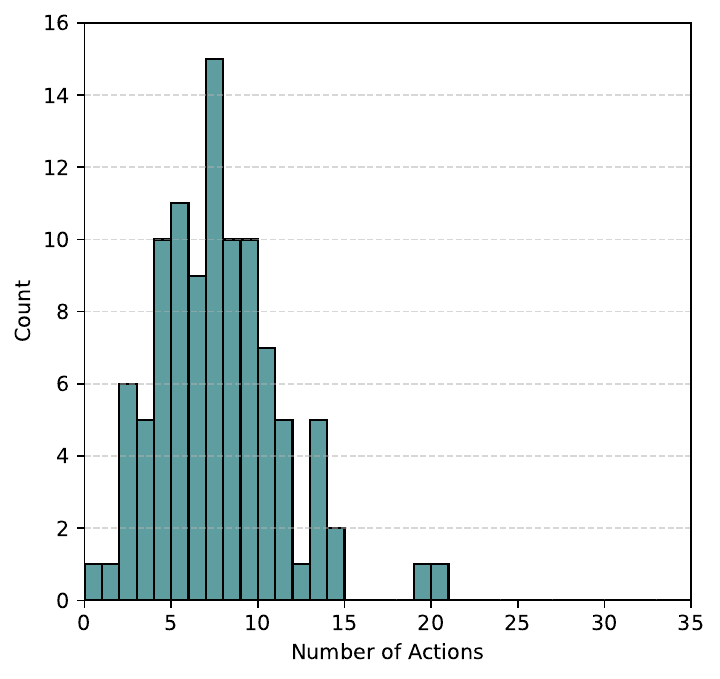}
  \captionof{figure}{Number of actions taken per episode for a sample of 100 episodes using a damped and stacked lateral navigation policy for 2 aircraft. A mean of 7.2 actions with standard deviation 3.5 is observed.}
  \label{fig:2_ac_lat_damp_stack_multi_run}
\end{minipage}
\end{figure}

To address the undesirable features of the undamped policy, we train a damped policy and apply our technique of \textit{online action-stacking}. The introduction of the action damping penalty described in Equation~\ref{sparse_action} significantly decreases the number of actions issued by the policy, as shown in Figure~\ref{fig:2_ac_damped_lat_example}. The policy now issues sparse commands, adjusting the aircraft's lateral profile only at key inflexion points along the route. Over a sample of 100 runs, the policy issues an average of 14.5 actions per episode, an 87\% reduction compared to the undamped policy. However, it can be seen that multiple commands are necessary to make larger turns, where operational ATC would replace this with a single compound instruction. The central turn made by the north-most aircraft near EGL with the damped policy consists of seven repeated "turn right 10 degrees" actions. This issue of operational realism is effectively addressed by our \textit{online action stacking} technique, as shown in Figure~\ref{fig:2_ac_stacked_lat_example}. Here, the central route turn is instead compiled into a single ``turn right 70 degrees'' command. Figures~\ref{fig:2_ac_lat_damp_multi_run}~and~\ref{fig:2_ac_lat_damp_stack_multi_run} demonstrate the significant reduction in action frequency caused by action stacking. A mean reduction in the number of actions per episode of \textasciitilde50\% is achieved across a sample of 100 episodes. Together, the action damping and action stacking reduce the number of issued actions to operationally relevant numbers.

\subsection{Lateral Navigation with Large Action Space}
\begin{figure}
\centering
\begin{minipage}[t]{.5\textwidth}
  \centering
  \captionsetup{width=.8\linewidth}
  \includegraphics[width=.9\linewidth]{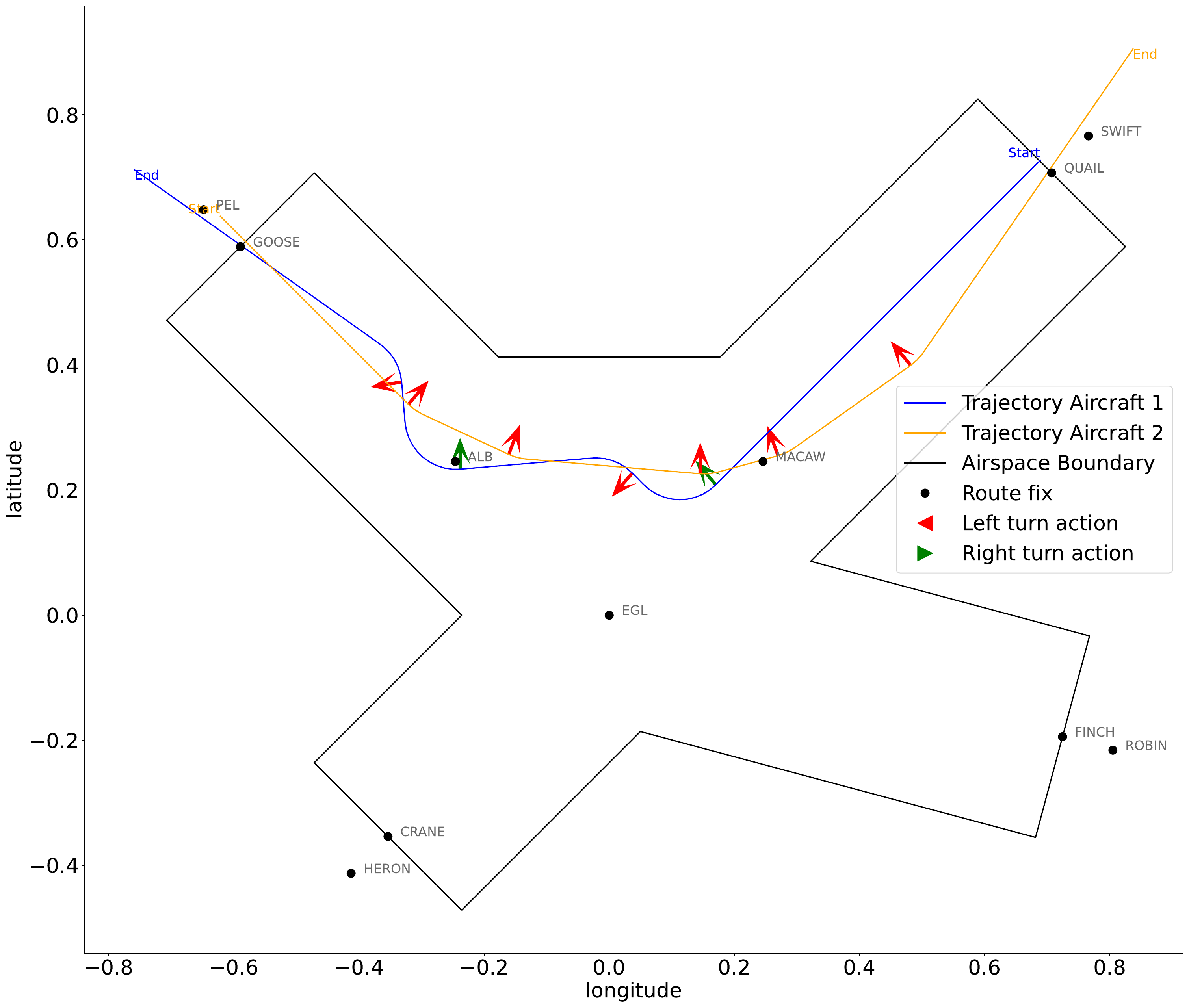}
  \captionof{figure}{Policy controlling 2 aircraft through the X-Plus sector with a broad action space.}
  \label{fig:big_act_2_ac_lat_example.pdf}
\end{minipage}%
\begin{minipage}[t]{.5\textwidth}
  \centering
  \captionsetup{width=.8\linewidth}
  \includegraphics[width=.94\linewidth]{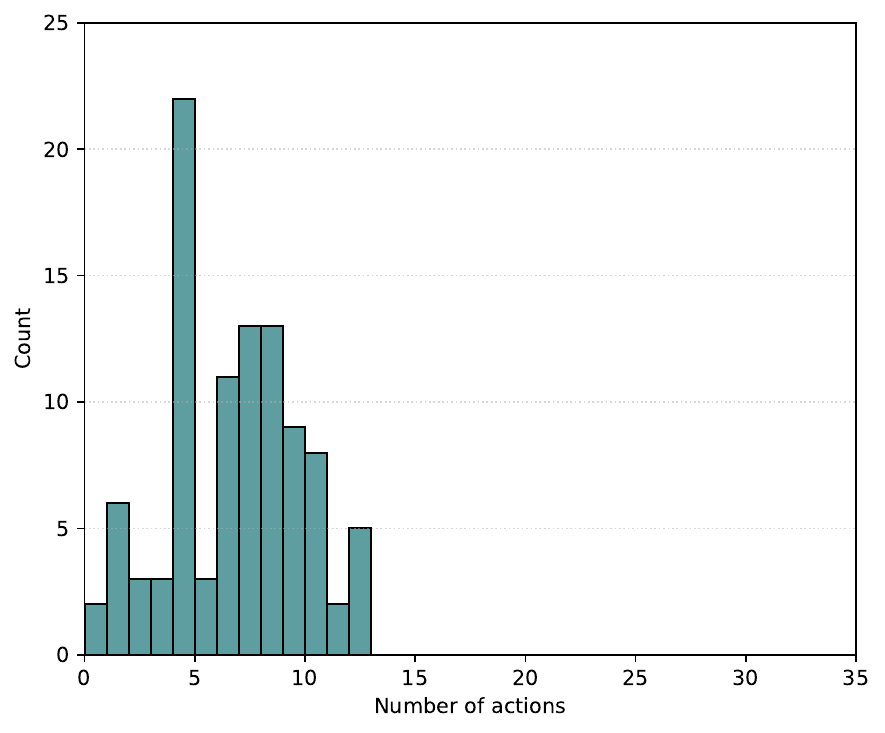}
  \captionof{figure}{Number of actions taken per episode for a sample of 100 episodes using a lateral navigation policy for 2 aircraft with a large action space. A mean of 6.3 actions with standard deviation 3.0 is observed.}
  \label{fig:big_act_multi_run}
\end{minipage}
\end{figure}

Figure~\ref{fig:big_act_2_ac_lat_example.pdf} shows an example of a policy trained with the larger lateral action space described in Section~\ref{sec:Action} (37 actions). After training for the same number of time steps, the policy learns to effectively navigate to the exit. However, as can be observed in the larger action space, this results in inefficient turns and over-correction. It is likely that this could be remedied through further training, but this is a key motivator for \textit{online action-stacking}: given the significant complexity of the full ATC task, reducing training time for comparable performance is a highly desirable feature. Figure~\ref{fig:big_act_multi_run} demonstrates that the policy issues a low average number of actions per episode, with a mean of 6.3 actions issued. Although the large action space policy achieves a lower mean action count (6.3 vs. 7.2), this difference is
not statistically significant (standard deviation $\approx 3$).
Crucially, the stacking approach achieves this comparable operation fidelity with significantly fewer actions available, reducing the ``curse of dimensionality'' often found in complex ATC formulations: the 5-dimensional action space performed similarly to a policy trained with a 37-dimensional action space.

\begin{figure}
\centering
\begin{minipage}[t]{.5\textwidth}
  \centering
  \captionsetup{width=.8\linewidth}
  \includegraphics[width=.9\linewidth]{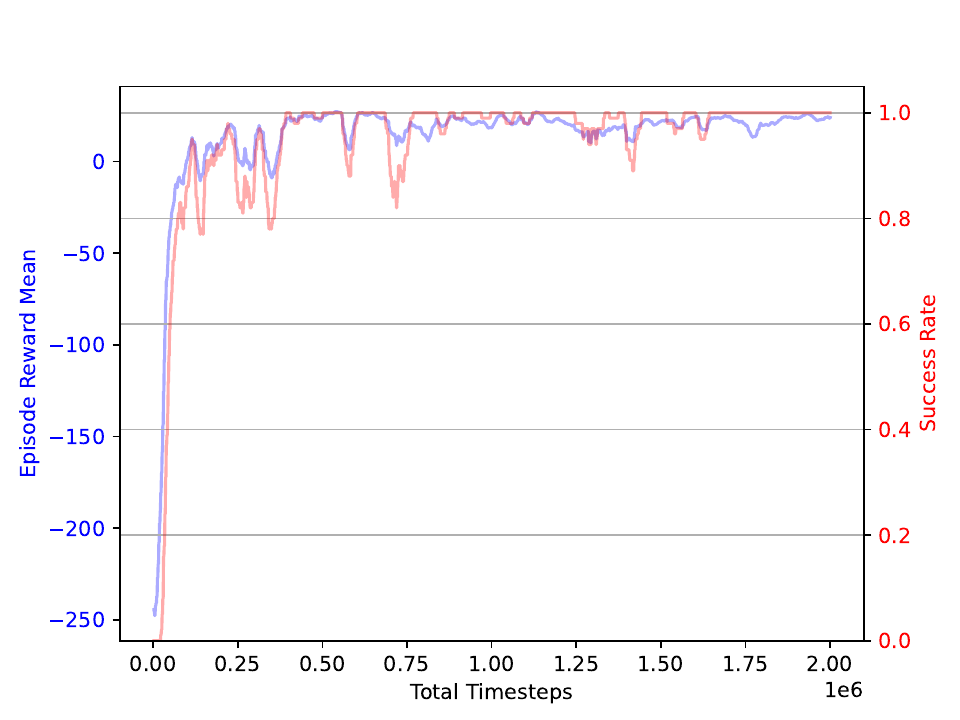}
  \captionof{figure}{Reward convergence for damped lateral navigation policy for 2 aircraft. Success is monitored using the success criteria defined in the terminal rewards in Section~\ref{sec:Reward}}
  \label{fig:stack_reward}
\end{minipage}%
\begin{minipage}[t]{.5\textwidth}
  \centering
  \captionsetup{width=.8\linewidth}
  \includegraphics[width=.9\linewidth]{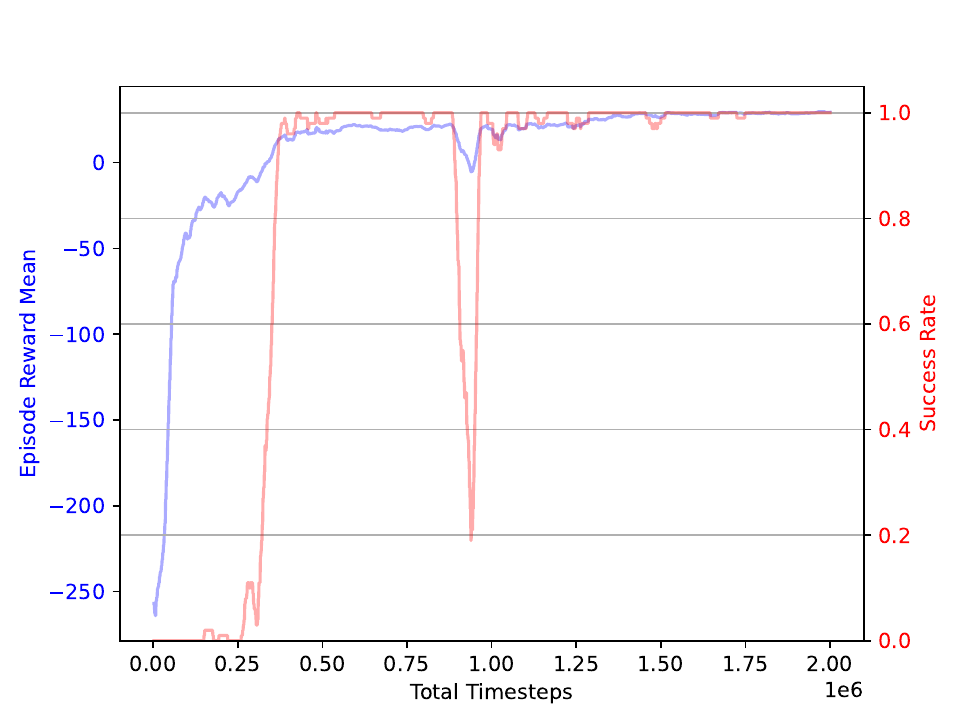}
  \captionof{figure}{Reward convergence for large action space lateral navigation policy for 2 aircraft. Success is monitored using the success criteria defined in the terminal rewards in Section~\ref{sec:Reward}}
  \label{fig:big_act_reward}
\end{minipage}
\end{figure}

Figures~\ref{fig:stack_reward}~and~\ref{fig:big_act_reward} show the %
rewards during training for both the damped lateral navigation policy and the navigation policy using the larger action space. It can be seen that the rate of convergence is indeed higher for the smaller action space, although, for this simpler problem, both policies train to a reasonable level of convergence at the same approximate level of reward. The success rate shows how often the agent satisfied all 3 of the terminal reward criteria described in Section~\ref{sec:Reward}: successful navigation to exit, no airspace excursions, and issuing fewer than 30 actions. It can be seen that reliable success occurs much later for the large action space policy. Inspection of the individual values reveals that it is the number-of-actions criterion
that caused the success rate drop for the large action space policy, despite the theoretical number of actions required to complete the scenario being smaller. Given the relative simplicity of this formulation of the ATC task, we anticipate significant challenges in scaling the approach when attempting more complex implementations with a broad action space.

These results suggest that it should be possible to approximately match the fidelity and efficacy of solutions with a smaller action space through the application of \textit{online action stacking}. Consequently, training complexity can be reallocated to handling higher traffic densities or more robust safety definitions without incurring prohibitive computational costs.

\subsection{Action Stacking for Vertical Control}

\begin{figure}[h]
\centering
\begin{minipage}{.5\textwidth}
  \centering
  \captionsetup{width=.8\linewidth}
  \includegraphics[width=.9\linewidth]{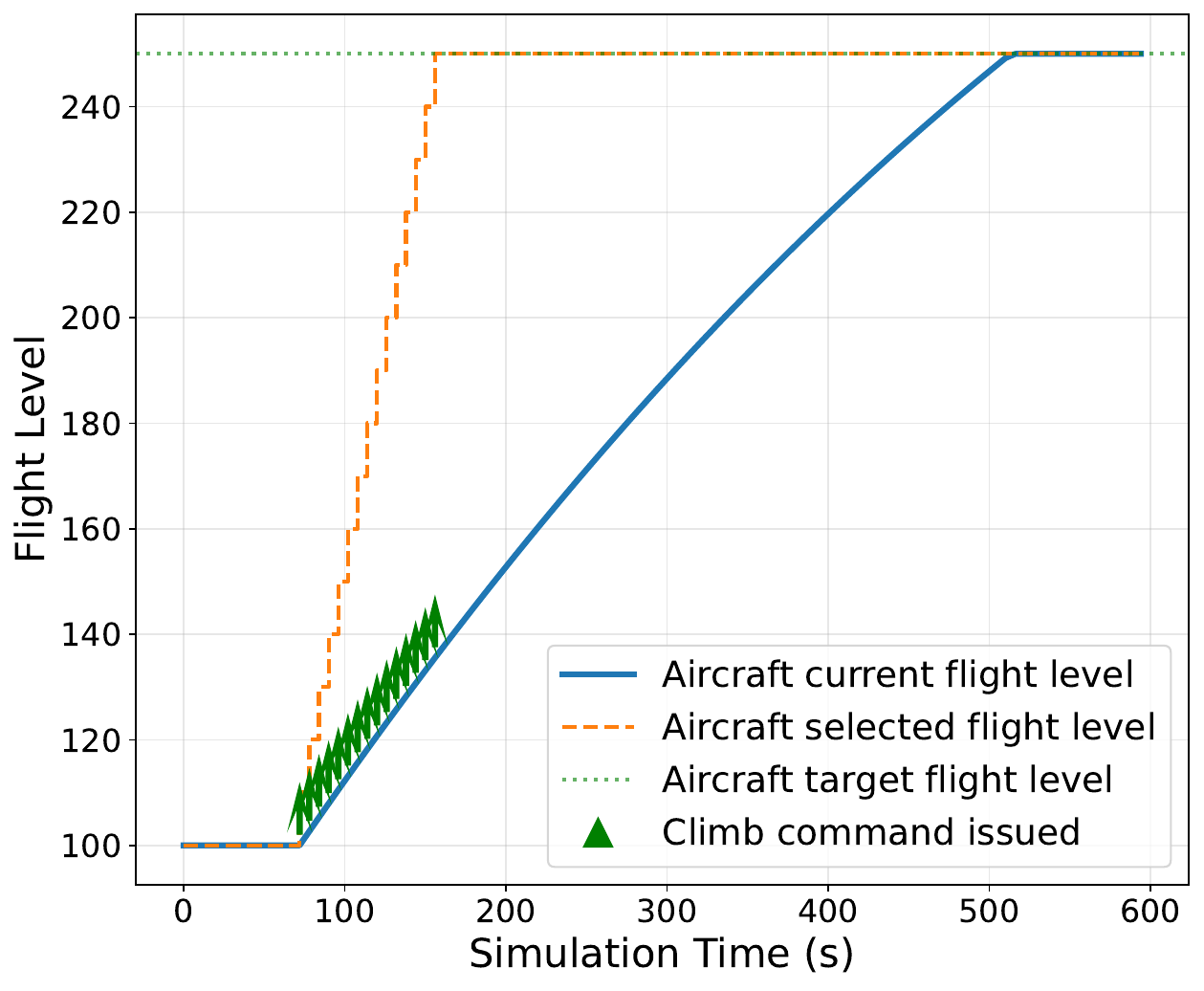}
  \captionof{figure}{Vertical control policy without stacking executing a climb. The sequence of control instructions can be seen increasing the selected level of the aircraft each simulation step.}
  \label{fig:vert_damped}
\end{minipage}%
\begin{minipage}{.5\textwidth}
  \centering
  \captionsetup{width=.8\linewidth}
  \includegraphics[width=.9\linewidth]{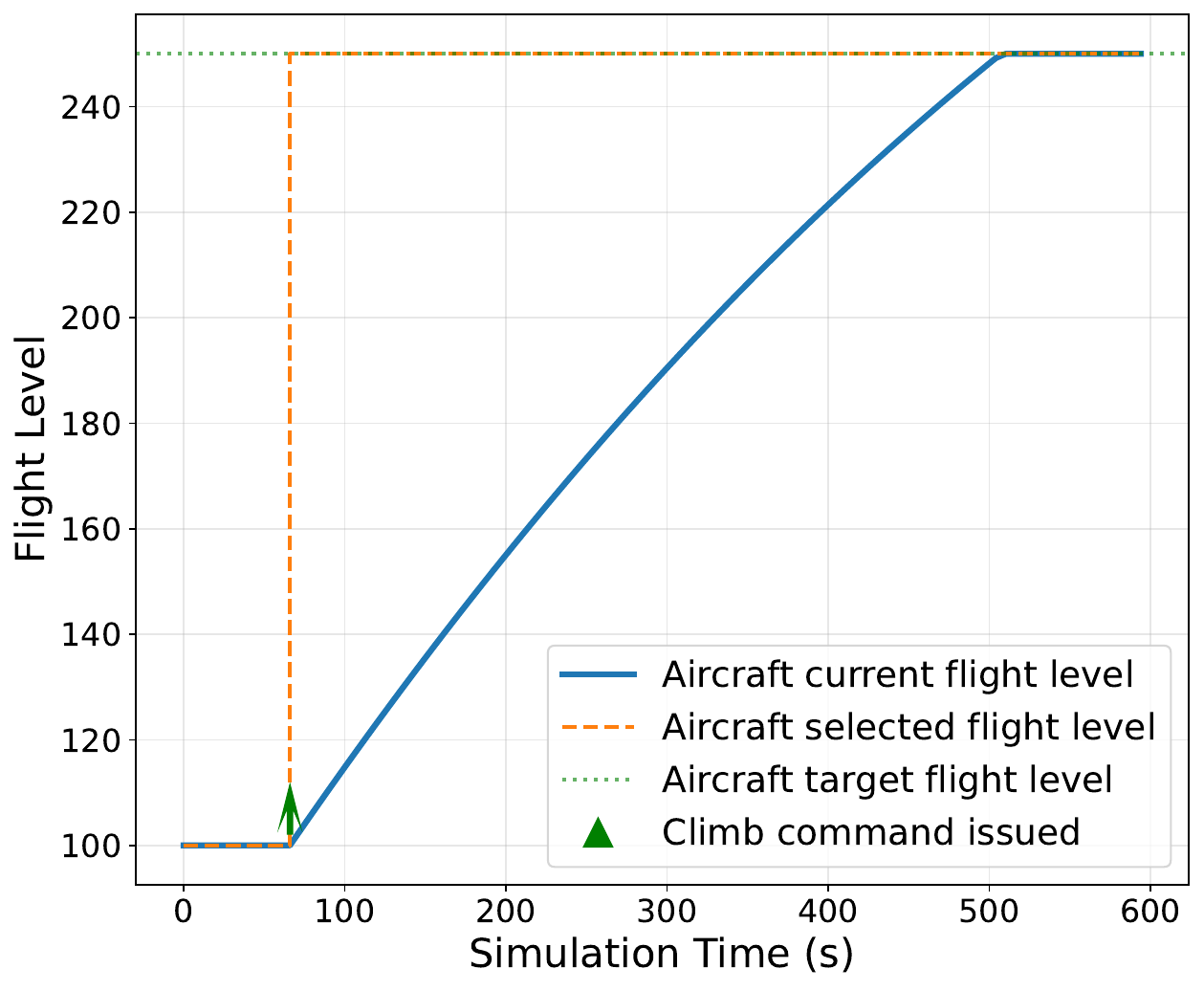}
  \captionof{figure}{Vertical control policy with action stacking demonstrating immediate clearance to target flight level upon action initiation. In this example 15 ``increase flight level'' instructions are compiled into a single command.}
  \label{fig:vert_stacked}
\end{minipage}
\end{figure}

Vertical control is a natural extension for action stacking because large changes in aircraft flight level are commonplace in ATC operations. To investigate this, a simple policy was trained to navigate aircraft to their target flight level, using the reward and state configuration discussed in the methodology and training for 2 million time steps. Figures~\ref{fig:vert_damped}~and~\ref{fig:vert_stacked} show the effect of applying action stacking in the vertical domain: a vertical instruction which would require 15 consecutive instructions to achieve is compiled into a single, operationally realistic climb command. Extension of \textit{online action stacking} to the vertical domain is of particular interest as it provides a solution to an otherwise highly challenging aspect of the ATC action space. Aircraft performance limits mean that the viable actions for each aircraft in the vertical will be different, as well as the issue of high dimensionality. A high performance jet aircraft may have an operational level range above flight level 400, resulting in a 40 dimensional action space just for a single aircraft as valid flight levels are multiples of 10. The application of \textit{online action stacking} has the potential to reduce this to only two dimensions.

\subsection{Action Stacking for Lateral Navigation and Avoidance}

\begin{figure}[h]
\centering
\begin{minipage}{.5\textwidth}
  \centering
  \captionsetup{width=.8\linewidth}
  \includegraphics[width=.9\linewidth]{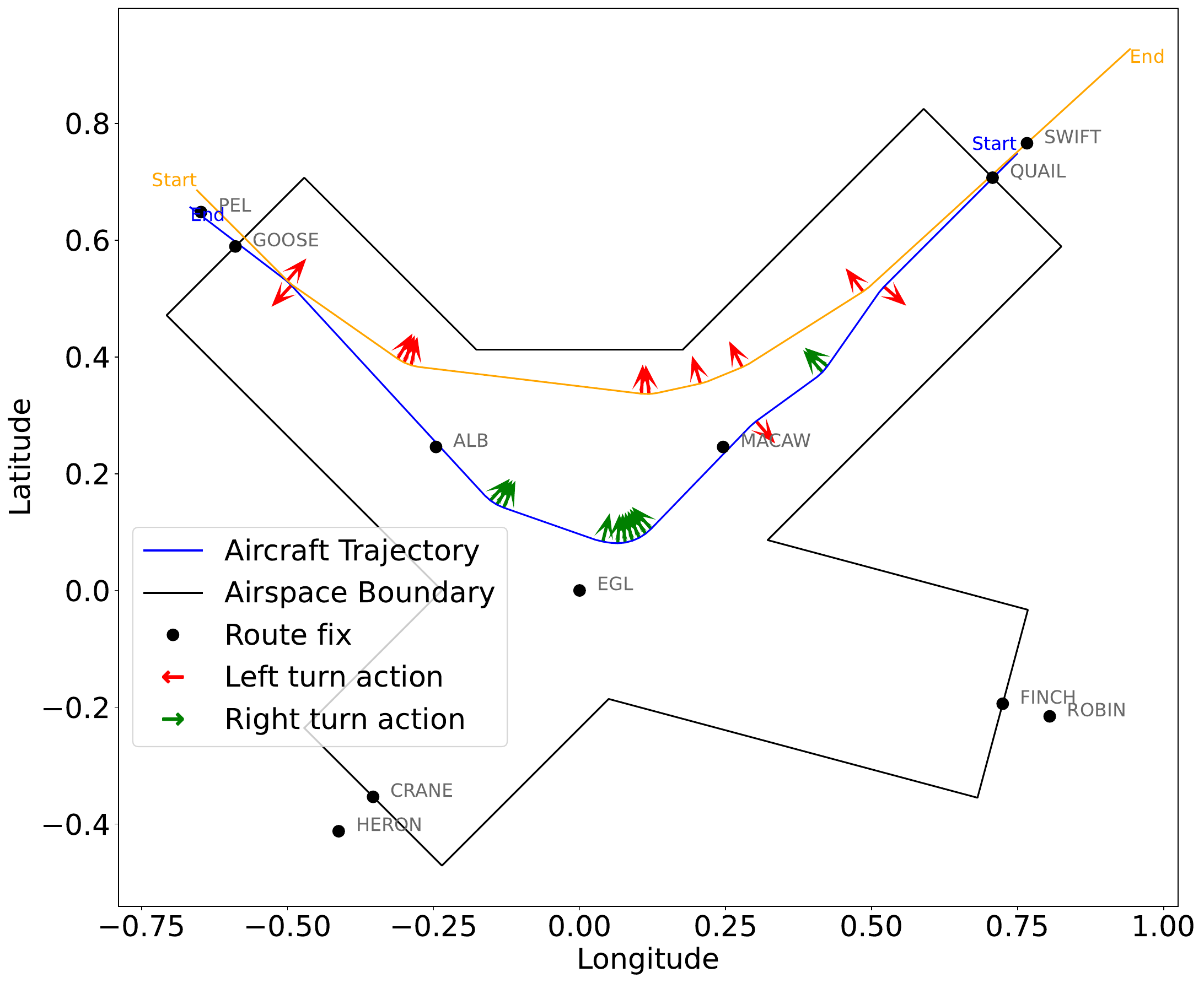}
  \captionof{figure}{Damped lateral navigation and avoidance policy controlling 2 aircraft.}
  \label{fig:avoid_damped}
\end{minipage}%
\begin{minipage}{.5\textwidth}
  \centering
  \captionsetup{width=.8\linewidth}
  \includegraphics[width=.9\linewidth]{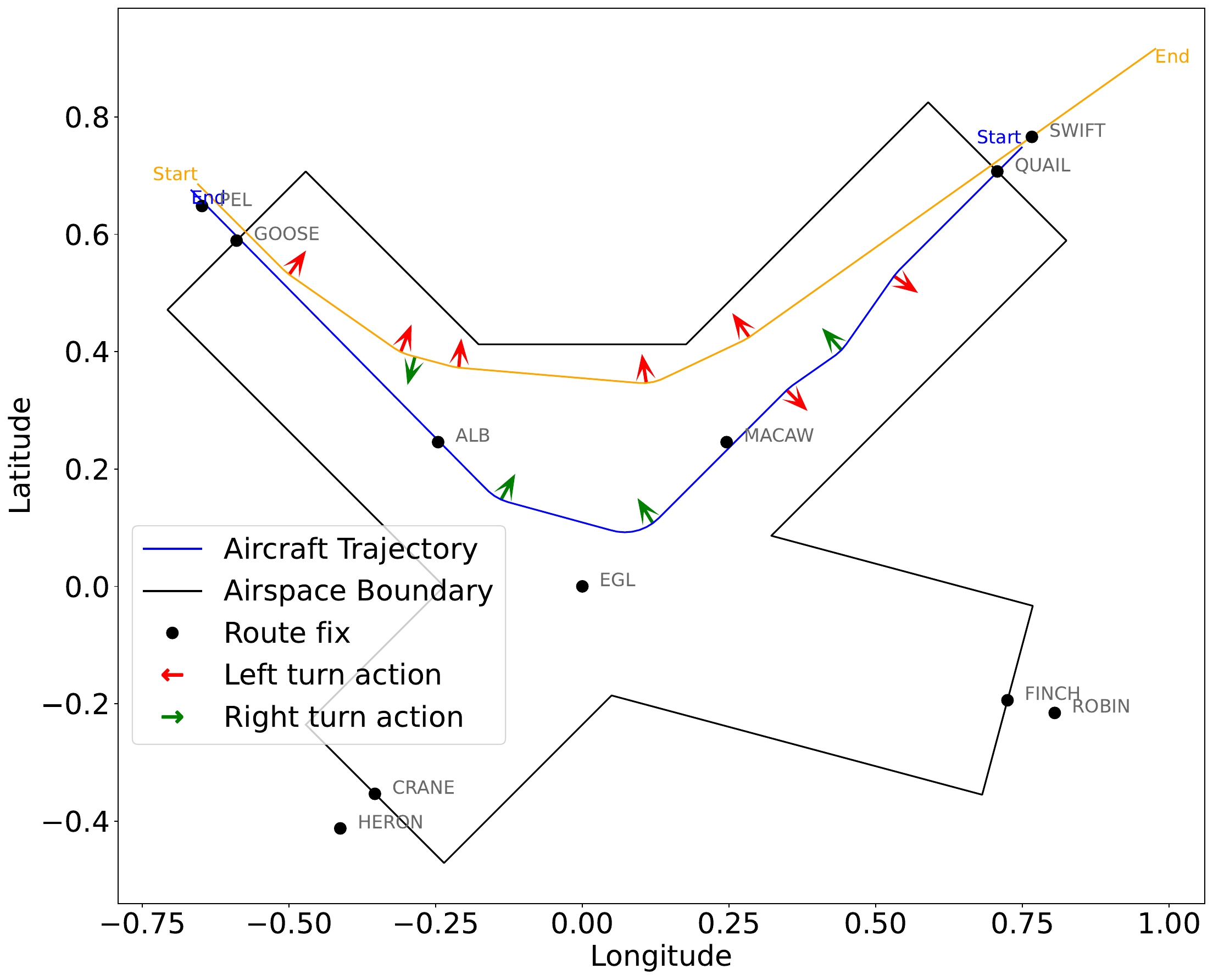}
  \captionof{figure}{Stacked lateral navigation and avoidance policy controlling 2 aircraft.}
  \label{fig:avoid_stacked}
\end{minipage}
\end{figure}

Adding in lateral avoidance behaviours significantly increases the complexity of the controlling task. The safety formulation described in Equation~\ref{eq:safety_reward} incentivises the ``fail-safe'' procedures that form the foundation of ATC. Figures~\ref{fig:avoid_damped}~and~\ref{fig:avoid_stacked} show damped and stacked policies that have been trained to provide collision avoidance, controlling two aircraft along reciprocal routes. It can be seen that, again, the central turns are effectively compiled in the stacked policy. In this example, 15 actions are issued by the damped policy, whereas only 7 are issued by the stacked policy, including two instructions of ``turn left 40 degrees'' and ``turn right 70 degrees''. Furthermore, effective separation is provided by vectoring one aircraft to the North, while bringing the other further South. A minimum separation of 18 nm is achieved in both scenarios, easily exceeding the 5 nm minimum.

\begin{figure}[h]
\centering
\begin{minipage}{.5\textwidth}
  \centering
  \captionsetup{width=.8\linewidth}
  \includegraphics[width=.9\linewidth]{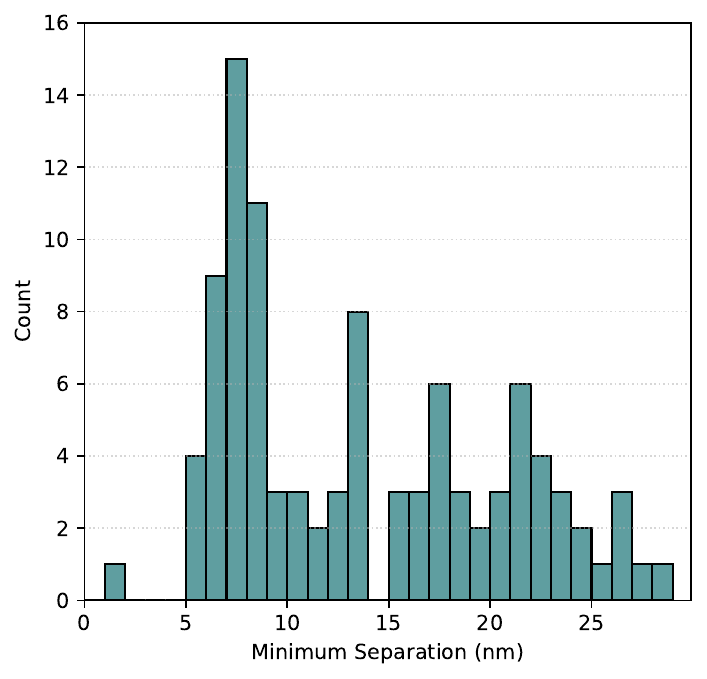}
  \captionof{figure}{Minimum separation achieved per episode for a stacked lateral navigation and avoidance policy.}
  \label{fig:avoid_sep_stacked}
\end{minipage}%
\begin{minipage}{.5\textwidth}
  \centering
  \captionsetup{width=.8\linewidth}
  \includegraphics[width=.9\linewidth]{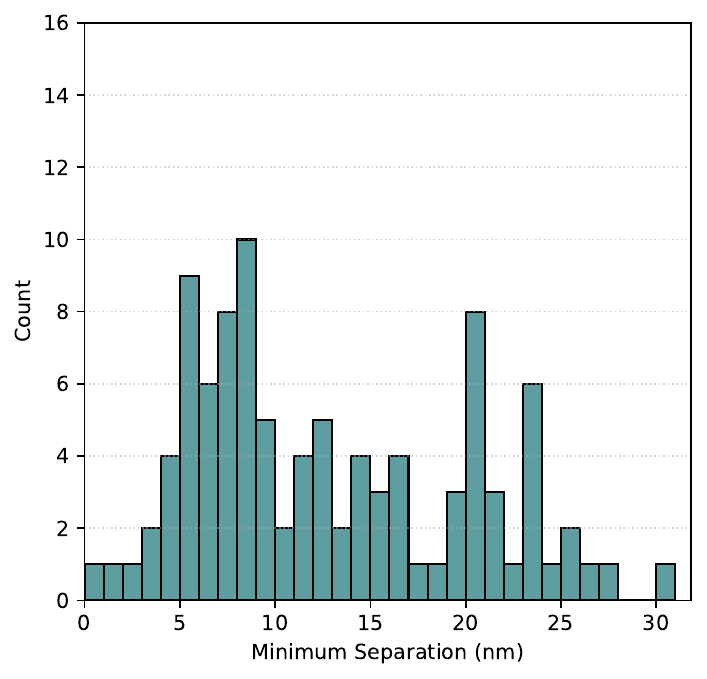}
  \captionof{figure}{Minimum separation achieved per episode for a large action space lateral navigation and avoidance policy.}
  \label{fig:avoid_sep_big_act}
\end{minipage}
\end{figure}

A core assertion of our motivation for online action-stacking is that more complex challenges such as lateral navigation and avoidance become harder to train when using a large action space. In addition to the lateral navigation and avoidance policy trained using our standard incremental action space described in Equation~\ref{eq:two_ac_act_space}, we therefore trained a second lateral navigation and avoidance policy using the large action space derived from Equation~\ref{2_ac_big_act_space} (19 actions per aircraft, 37 actions in total for two aircraft). Figures~\ref{fig:avoid_sep_stacked}~and~\ref{fig:avoid_sep_big_act} show the minimum separation achieved per episode for a sample of 100 episodes for each of these policies. Only one loss of separation (less than 5 nm) occurred for the stacked policy, whereas the large action space policy incurred nine losses of separation. Furthermore, we report a mean number of actions per episode for the stacked policy of 81.2 with a standard deviation of 13.2, while the large action space policy issued a mean of 126.3 actions per episode with a standard deviation of 12.3. This directly supports our assertion that smaller action spaces yield more effective training for complex ATC tasks under a fixed training budget, as both action sparsity and separation performance are superior in the damped/stacked policy. Both policies were trained for more than 30 million steps (around 60 hours on a modern server), so while additional training is always possible, we gain a clear advantage if we can use a smaller action space without loss of fidelity.

\section{Conclusion}

We have developed a novel approach that combines online action-stacking with the incentivisation of burst actions to achieve action sparsity, replicate the compound instructions of much larger action spaces, and directly address the reinforcement learning training challenges posed by ATC. This technique enables realistic controller-style commands while maintaining training efficiency. For lateral navigation, we have shown that comparable results can be achieved using online action-stacking with a 5-dimensional action space instead of a 37-dimensional action space.

Our current work on two-aircraft avoidance and navigation demonstrates promising results. In future work, we will extend our application to combine both lateral and vertical control with deconfliction, thus providing a comprehensive treatment of the fundamental ATC task and addressing a significant gap in the existing literature.

\section*{Acknowledgments}
This work was supported by the EPSRC (EP/V056522/1) and NATS.

\bibliography{references_modified}

\end{document}